\documentclass[fleqn,10pt]{wlscirep}
\usepackage[utf8]{inputenc}
\usepackage[T1]{fontenc}
\title{Fracture Detection in Pediatric Wrist Trauma X-ray Images Using YOLOv8 Algorithm}

\author[1]{Rui-Yang Ju}
\author[2,*]{Weiming Cai}
\affil[1]{National Taiwan University, Graduate Institute of Networking and Multimedia, Taipei City, 106335, Taiwan}
\affil[2]{Jingjiang People's Hospital, Department of Hand and Foot Surgery, Jingjiang City, 214500, China}
\affil[*]{Corresponding author: 1318746637@qq.com}


\begin{abstract}
Hospital emergency departments frequently receive lots of bone fracture cases, with pediatric wrist trauma fracture accounting for the majority of them. Before pediatric surgeons perform surgery, they need to ask patients how the fracture occurred and analyze the fracture situation by interpreting X-ray images. The interpretation of X-ray images often requires a combination of techniques from radiologists and surgeons, which requires time-consuming specialized training. With the rise of deep learning in the field of computer vision, network models applying for fracture detection has become an important research topic. In this paper, we use data augmentation to improve the model performance of YOLOv8 algorithm (the latest version of You Only Look Once) on a pediatric wrist trauma X-ray dataset (GRAZPEDWRI-DX), which is a public dataset. The experimental results show that our model has reached the state-of-the-art (SOTA) mean average precision (mAP 50). Specifically, mAP 50 of our model is 0.638, which is significantly higher than the 0.634 and 0.636 of the improved YOLOv7 and original YOLOv8 models. To enable surgeons to use our model for fracture detection on pediatric wrist trauma X-ray images, we have designed the application ``Fracture Detection Using YOLOv8 App'' to assist surgeons in diagnosing fractures, reducing the probability of error analysis, and providing more useful information for surgery.
\end{abstract}

\begin{document}
\flushbottom
\maketitle
\thispagestyle{empty}

\section*{Introduction}
In hospital emergency rooms, radiologists are often asked to examine patients with fractures in various parts of the body, such as the wrist and arm. Fractures can generally be classified as open or closed, with open fractures occurring when the bone pierces the skin, and closed fractures occurring when the skin remains intact despite the broken bone. Before performing surgery, the surgeon must inquire about the medical history of the patients and conduct a thorough examination to diagnose fracture. In recent medical imaging, three types of devices, including X-ray, Magnetic Resonance Imaging (MRI), and Computed Tomography (CT), are commonly used to diagnose fracture \cite{hopkins2021}. And X-ray is the most widely used device due to its cost-effectiveness.

Fractures of the distal radius and ulna account for the majority of wrist trauma in pediatric patients \cite{hedstrom2010epidemiology,randsborg2013fractures}. In prestigious hospitals of developed areas, there are many experienced radiologists who are capable of correctly analyzing X-ray images; while in some small hospitals of underdeveloped regions, there are only young and inexperienced surgeons who may be unable to correctly interpret X-ray images. Therefore, a shortage of radiologists would seriously jeopardize timely patient care \cite{burki2018shortfall,rimmer2017radiologist}. Specifically, some hospitals in Africa have even limited access to specialist reports \cite{rosman2015imaging}, which badly affects the probaility of the sucess of surgery. According to the survey \cite{mounts2011most,erhan2013overlooked}, the percentage of X-ray images misinterpreted have reached 26\%.

With the advancement of deep learning, neural network models have been introduced in medical image processing \cite{adams2021artificial,tanzi2020hierarchical,chung2018automated,choi2020using}. In recent years, researchers have started to apply object detection models \cite{girshick2014rich,ju2022threshnet,ju2023efficient} to fracture detection \cite{gan2019artificial,kim2018artificial,lindsey2018deep,bluthgen2020detection}, which is a popular research topic in computer vision (CV).

Deep learning methods in the field of object detection are divided into two-stage and one-stage algorithms. Two-stage algorithm models such as R-CNN \cite{girshick2014rich} and its improved models \cite{girshick2015fast,ren2015faster,cai2018cascade,lu2019grid,pang2019libra,zhang2020dynamic,he2017mask} generate location and class probabilities in two stages. Whereas one-stage algorithm models directly produce the location and class probabilities of objects, resulting in the improvement of the model inference speed. In addition to the classical one-stage algorithm models, such as SSD \cite{liu2016ssd}, RetinaNet \cite{lin2017focal}, CornerNet \cite{law2018cornernet}, CenterNet \cite{duan2019centernet}, and CentripetalNet \cite{dong2020centripetalnet}, You Only Look Once (YOLO) series algorithm models \cite{wang2022yolov7,ge2021yolox,wang2021yolor} are preferred for real-time applications \cite{ju2023resolution} due to the good balance between the model accuracy and inference speed.

In this paper, we first use YOLOv8 algorithm \cite{glenn2023} to train models of different sizes on the GRAZPEDWRI-DX \cite{nagy2022pediatric} dataset. After evaluation of the model performances of YOLOv8, we train the models by using data augmentation to detect wrist fractures in children. We compare YOLOv8 models using our training method with YOLOv7 and its improved models, and the experimental results demonstrate that our models have the highest the mean average precision (mAP 50) value.

The contributions of this paper are summarized as follows:
\begin{itemize}
\item
We use data augmentation to improve the model performance of YOLOv8 model. The experimental results show that the mean average precision of YOLOv8 model using our training method for fracture detection on the GRAZPEDWRI-DX dataset reaches SOTA value.
\item
This work develops an application to detect wrist fracture in children, which aims to help pediatric surgeons interpret X-ray images without the assistance of the radiologist, and reduce the probability of X-ray image analysis errors.
\end{itemize}

This paper is structured as follows: Section \textbf{Related Work} describes the deep learning methods for detecting fracture, and describes the application of YOLOv5 model in medical image processing. Section \textbf{Proposed Method} introduces the whole process of training and the architecture of our model. Section \textbf{Experiments} presents the improved performance of YOLOv8 model using our training method compared with YOLOv7 and its improved models. Section \textbf{Application} describes our proposed application to assist pediatric surgeons in analyzing X-ray images. Finally, Section \textbf{Conclusions and Future Work} discusses the conclusions and future work of this paper.

\begin{table}[ht]
\caption{Experimental results of other studies on fracture detection in various parts of the body based on deep learning method.}
\label{other}
\begin{tabular}{ccccc}
\hline
\textbf{Author} & \textbf{Task} & \textbf{Model} & \textbf{Dataset} & \textbf{mAP$\rm ^{val}$ 50} \\
\hline
Guan \emph{et al.} \cite{guan2019thigh} & Thigh Fracture Detection & DCFPN & 3,842 thigh fracture X-ray radiographs & 0.821 \\
Wang \emph{et al.} \cite{wang2021parallelnet} & Thigh Fracture Detection & R-CNN & 3,842 thigh fracture X-ray radiographs & 0.878 \\
Guan \emph{et al.} \cite{guan2020arm} & Arm Fracture Detection & R-CNN & Musculoskeletal-Radiograph (MURA) \cite{rajpurkar2018mura} & 0.620 \\
Wu \emph{et al.} \cite{wu2021feature} & Bone Fracture Detection & FAMO & 9,040 radiographs of various body parts & 0.774 \\
Ma and Luo \cite{ma2021bone} & Bone Fracture Detection & Faster R-CNN & 1,052 bone x-ray images & 0.884 \\
Xue \emph{et al.} \cite{xue2021detection} & Hand Fracture Detection & Faster R-CNN & 3,067 hand trauma x-ray images & 0.700 \\
Sha \emph{et al.} \cite{sha2020detectionyolo} & Spine Fracture Detection & Faster R-CNN & 5,134 spine fractures CT images & 0.733 \\
Sha \emph{et al.} \cite{sha2020detectionrcnn} & Spine Fracture Detection & YOLOv2 & 5,134 spine fractures CT images & 0.753 \\ \hline
\end{tabular}
\end{table}

\section*{Related Work}
In recent years, neural networks have been widely utilized in image data for fracture detection. Guan \emph{et al.} \cite{guan2019thigh} achieved the average precision of 82.1\% on 3,842 thigh fracture X-ray images using the Dilated Convolutional Feature Pyramid Network (DCFPN). Wang \emph{et al.} \cite{wang2021parallelnet} employed a novel R-CNN \cite{girshick2014rich} network ParalleNet as the backbone network for fracture detection on 3,842 thigh fracture X-ray images. In addition to thigh fracture detection, about arm fracture detection, Guan \emph{et al.} \cite{guan2020arm} used R-CNN for detection on Musculoskeletal-Radiograph (MURA) dataset \cite{rajpurkar2018mura} and obtained an average precision of 62.04\%. Ma and Luo \cite{ma2021bone} used Faster R-CNN \cite{ren2015faster} for fracture detection on a part of 1,052 bone images of the dataset and the proposed CrackNet model for fracture classification on the whole dataset. Wu \emph{et al.} \cite{wu2021feature} proposed Feature Ambiguity Mitigate Operator (FAMO) model based on ResNeXt101 \cite{xie2017aggregated} and FPN \cite{lin2017feature} for bone fracture detection on 9,040 radiographs of various body parts. Qi \emph{et al.} \cite{qi2020ground} utilized Fast R-CNN \cite{girshick2015fast} with ResNet50 \cite{he2016deep} as the backbone network to detect nine different types of fractures on 2,333 fracture X-ray images. Xue \emph{et al.} \cite{xue2021detection} utilized the Faster R-CNN model for hand fracture detection on 3,067 hand trauma X-ray images, achieving an average precision of 70.0\%. Sha \emph{et al.} \cite{sha2020detectionyolo,sha2020detectionrcnn} used YOLOv2 \cite{redmon2017yolo9000} and Faster R-CNN \cite{ren2015faster} models for fracture detection on 5,134 CT images of spine fractures respectively. Experiments showed that the average precision of YOLOv2 reached 75.3\%, which was higher than 73.3\% of Faster R-CNN, and inference time of YOLOv2 for each CT image is 27ms, which is much faster than 381ms of Faster R-CNN. From Table \ref{other}, it can be seen that even though most of the works using R-CNN series models have shown excellent results, the inference speed is not satisfactory.

\begin{figure}[ht]
  \centering
  \includegraphics[width=\linewidth]{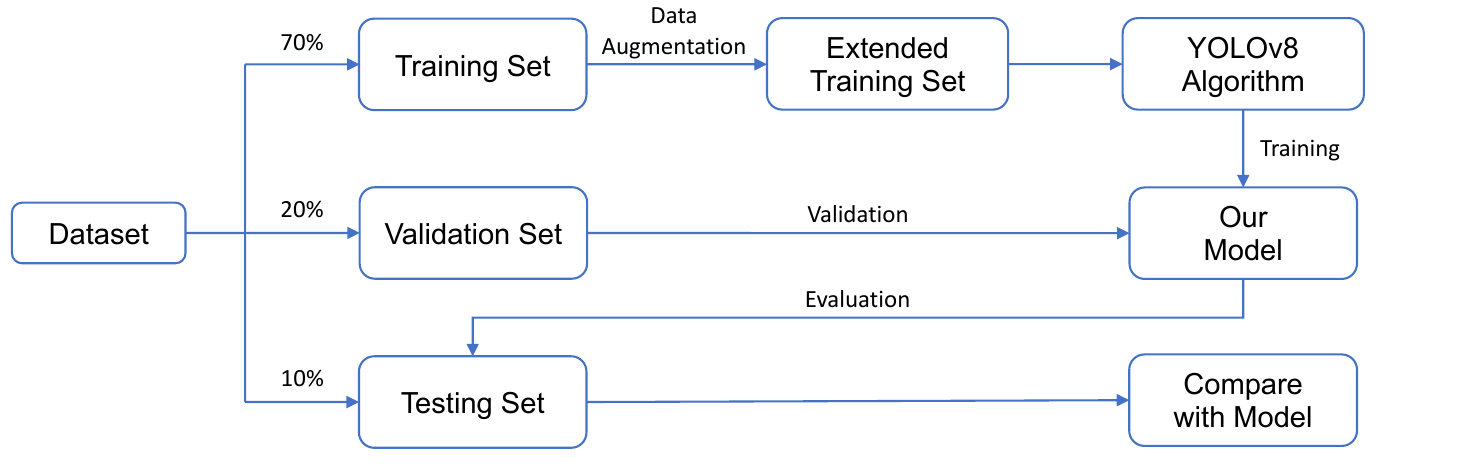}
  \caption{Flowchart of the model training, validation and testing on the dataset. The extended training set is used to double the number of X-ray images by data augmentation.}
  \label{fig_overall}
\end{figure}

\begin{figure}[ht]
  \centering
  \includegraphics[width=\linewidth]{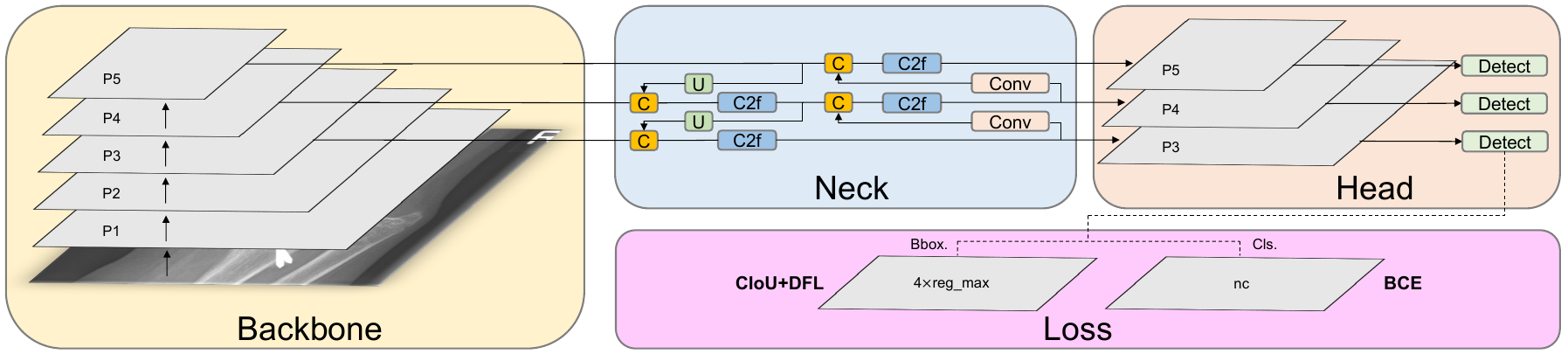}
  \caption{The architecture of YOLOv8 algorithm, which is divided into four parts, including backbone, neck, head, and loss.}
  \label{fig_arch}
\end{figure}

YOLO series models \cite{wang2022yolov7,ge2021yolox,wang2021yolor} offer a balance of performance in terms of the model accuracy and inference speed, which is suitable for mobile devices in real-time X-ray images detection. Hržić \emph{et al.} \cite{hrvzic2022fracture} proposed a machine learning model based on YOLOv4 method to help radiologists diagnose fractures and demonstrated that the AUC-ROC (area under the receiver operator characteristic curve) value of YOLO 512 Anchor model-AI was significantly higher than that of radiologists. YOLOv5 model \cite{glenn2022}, which was proposed by Ultralytics in 2021, has been deployed on mobile phones as the ``iDetection'' application. On this basis, Yuan \emph{et al.} \cite{yuan2022improved} employed external attention and 3D feature fusion techniques in YOLOv5 model to detect skull fractures in CT images. Warin \emph{et al.} \cite{warin2023maxillofacial} used YOLOv5 model to detect maxillofacial fractures in 3,407 maxillofacial bone CT images, and classified the fracture conditions into frontal, midfacial, mandibular fractures and no fracture. Rib fractures are a precursor injury to physical abuse in children, and chest X-ray (CXR) images are preferred for effective diagnosis of rib fracture conditions because of their convenience and low radiation dose. Tsai \emph{et al.} \cite{tsai2022automatic} used data augmentation with YOLOv5 model to detect rib fractures in CXR images. And Burkow \emph{et al.} \cite{burkow2022avalanche} applied YOLOv5 model to detect rib fractures in 704 pediatric CXR images, the model obtained the F2 score value of 0.58. To identify and detect mandibular fractures in panoramic radiographs, Warin \emph{et al.} \cite{warin2022assessment} used convolutional neural networks (CNNs) and YOLOv5 model to implement it. Fatima \emph{et al.} \cite{fatima2022vertebrae} used YOLOv5 model to localize vertebrae, which is important for detecting spinal deformities and fractures, and obtained an average precision of 0.94 at an IoU (Intersection over Union) threshold of 0.5. Moreover, Mushtaq \emph{et al.} \cite{mushtaq2022localization} applied YOLOv5 model to localize the lumbar spine and obtained an average precision value of 0.975. Nevertheless, relatively few researches have been reported on pediatric wrist fracture detection using YOLOv5 model. While YOLOv8 was proposed by Ultralytics in 2023, we use this algorithm to train the model for the first time in pediatric wrist fracture detection.

\section*{Proposed Method}
In this section, we introduce the process of the model training, validation and testing on the dataset, the architecture of YOLOv8 model, and the data augmentation technique employed during training. Figure \ref{fig_overall} illustrates the flowchart depicting the model training process and performance evaluation. We randomly divide the 20,327 X-ray images of the GRAZPEDWRI-DX dataset into the training, validation, and test set, where the training set is expanded to 28,408 X-ray images by data augmentation from the original 14,204 X-ray images. We design our model according to YOLOv8 algorithm, and the architecture of YOLOv8 algorithm is shown in Figure \ref{fig_arch}.

\subsection*{Data Augmentation}
\begin{figure}[t]
  \centering
  \includegraphics[width=\linewidth]{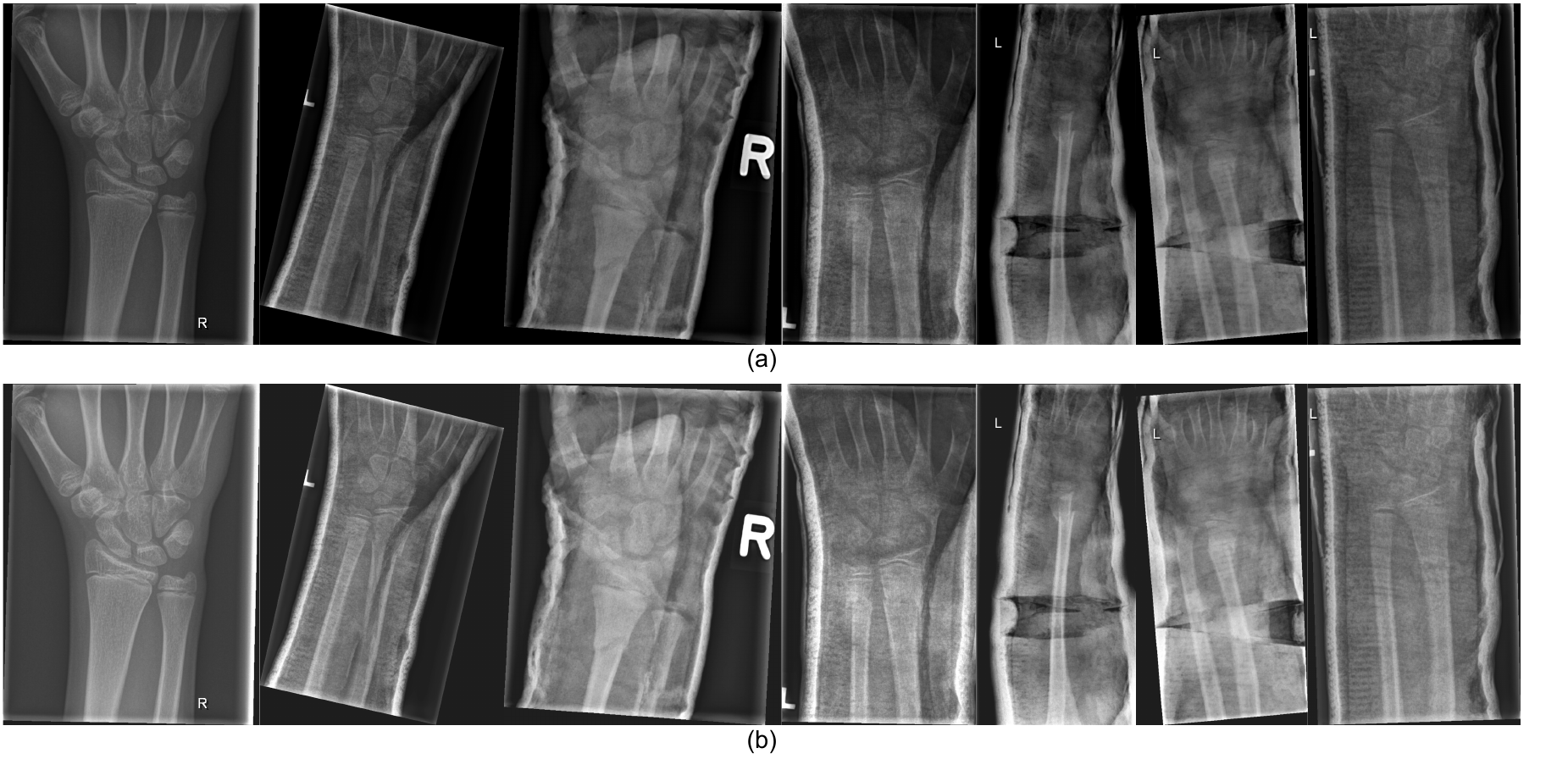}
  \caption{Examples of pediatric wrist X-ray images using data augmentation. (a) the original images, (b) the adjusted images.}
  \label{fig_aug}
\end{figure}

During the model training process, data augmentation is employed in this work to extend the dataset. Specifically, we adjust the contrast and brightness of the original X-ray image to enhance the visibility of bone-anomaly. This is achieved using the addWeighted function available in OpenCV (Open Source Computer Vision Library). The equation is presented below:
\begin{equation}
Output = Input_1 \times \alpha + Input_2 \times \beta + \gamma,
\end{equation}
where $Input_1$ and $Input_2$ are the two input images of the same size respectively, $\alpha$ represents the weight assigned to the first input image, $\beta$ denotes the weight assigned to the second input image, and $\gamma$ represents the scalar value added to each sum. Since our purpose is to adjust the contrast and brightness of the original input image, we take the same image as $Input_1$ and $Input_2$ respectively and set $\beta$ to 0. The value of $\alpha$ and $\gamma$ represent the proportion of the contrast and the brightness of the image respectively. The image after adjusting the contrast and brightness is shown in Figure \ref{fig_aug}. After comparing different settings, we finally decided to set $\alpha$ to 1.2 and $\gamma$ to 30 to avoid the output image being too bright.

\begin{figure}[ht]
  \centering
  \includegraphics[width=\linewidth]{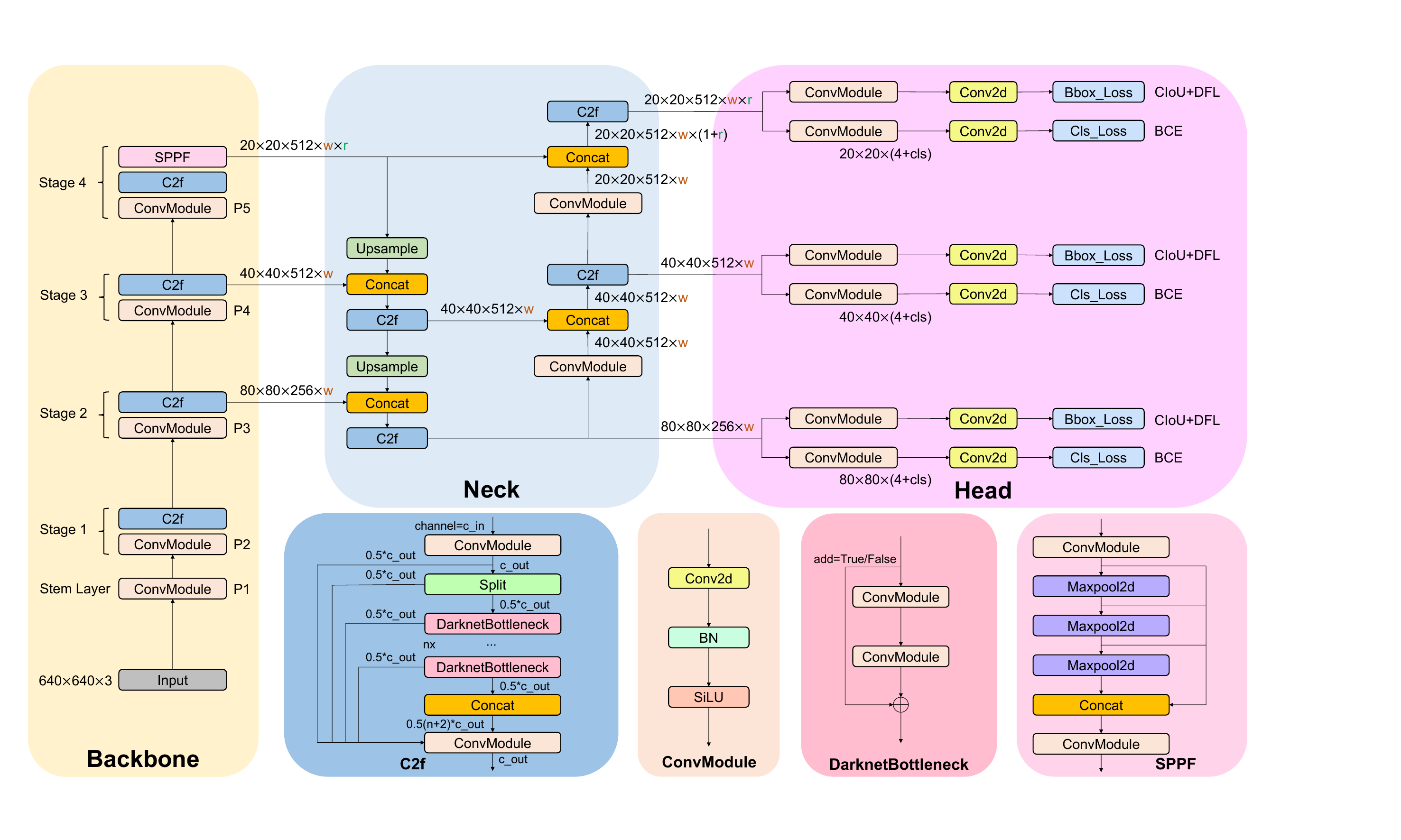}
  \caption{Detailed illustration of YOLOv8 model architecture. The Backbone, Neck, and Head are the three parts of our model, and C2f, ConvModule, DarknetBottleneck, and SPPF are modules.}
  \label{fig_detail}
\end{figure}

\subsection*{Model Architecture}
Our model architecture consists of backbone, neck, and head, as shown in Figure \ref{fig_detail}.  In the following subsections, we introduce the design concepts of each part of the model architecture, and the modules of different parts.

\subsubsection*{Backbone}
The backbone of the model uses Cross Stage Partial (CSP) \cite{wang2020cspnet} architecture to split the feature map into two parts. The first part uses convolution operations, and the second part is concatenated with the output of the previous part. The CSP architecture improves the learning ability of the CNNs and reduces the computational cost of the model.

YOLOv8 \cite{glenn2023} introduces C2f module by combining the C3 module and the concept of ELAN from YOLOv7 \cite{wang2022yolov7}, which allows the model to obtain richer gradient flow information. The C3 module consists of 3 ConvModule and \emph{n} DarknetBottleNeck, and the C2f module consists of 2 ConvModule and \emph{n} DarknetBottleNeck connected through Split and Concat, as illustrated in Figure \ref{fig_detail},  where the ConvModule consists of \emph{Conv-BN-SiLU}, and \emph{n} is the number of the bottleneck. Unlike YOLOv5 \cite{glenn2022}, we use the C2f module instead of the C3 module.

Furthermore, we reduce the number of blocks in each stage compared to YOLOv5 to further reduce the computational cost. Specifically, our model reduces the number of blocks to 3,6,6,3 in Stage 1 to Stage 4, respectively. Additionally, we adopt the Spatial Pyramid Pooling - Fast (SPPF) module in Stage 4, which is an improvement from Spatial Pyramid Pooling (SPP) \cite{he2015spatial} to improve the inference speed of the model. These modifications lead to our model with a better learning ability and shorter inference time.

\subsubsection*{Neck}
Generally, deeper networks obtain more feature information, resulting in better dense prediction. However, excessively deep networks reduce the location information of the object, and too many convolution operations will lead to information loss for small objects. Therefore, it is necessary to use Feature Pyramid Network (FPN) \cite{lin2017feature} and Path Aggregation Network (PAN) \cite{liu2018path} architectures for multi-scale feature fusion. As illustrated in Figure \ref{fig_detail}, the Neck part of our model architecture uses multi-scale feature fusion to combine features from different layers of the network. The upper layers acquire more information due to the additional network layers, whereas the lower layers preserve location information due to fewer convolution layers.

Inspired by YOLOv5, where FPN upsamples from top to bottom to increase the amount of feature information in the bottom feature map; and PAN downsamples from bottom to top to obtain more the top feature map information. These two feature outputs are merged to ensure precise predictions for images of various sizes. We adopt FP-PAN (Feature Pyramid-Path Aggregation Network) in our model, and delete convolution operations in upsampling to reduce the computational cost.

\subsubsection*{Head}
Different from YOLOv5 model utilizing a coupled head, we use a decoupled head \cite{ge2021yolox}, where the classification and detection heads are separated. Figure \ref{fig_detail} illustrates that our model deletes the objectness branch and only retains the classification and regression branches. Anchor-Base employes a large number of anchors in the image to determine the four offsets of the regression object from the anchors. It adjusts the precise object location using the corresponding anchors and offsets. In contrast, we adopt Anchor-Free \cite{tian2020fcos}, which identifies the center of the object and estimates the distance between the center and the bounding box.

\subsubsection*{Loss}
For positive and negative sample assignment, the Task Aligned Assigner of Task-aligned One-stage Object Detection (TOOD) \cite{feng2021tood} is used in our model training to select positive samples based on the weighted scores of classification and regression, as shown in Equation \ref{tood} below:
\begin{equation}
\label{tood}
t=s^\alpha \times u^\beta,
\end{equation}
where $s$ is the predicted score corresponding to the labeled class, and $u$ is the IoU of the prediction and the ground truth bounding box.

In addition, our model has classification and regression branches, where the classification branch uses Binary Cross-Entropy (BCE) Loss, and the equation is shown below:
\begin{equation}
\label{bce}
Loss_{n}=-w\left[y_{n} \log x_{n}+\left(1-y_{n}\right) \log \left(1-x_{n}\right)\right],
\end{equation}
where $w$ is the weight, $y_n$ is the labeled value, and $x_n$ is the predicted value of the model. 

The regression branch uses Distribute Focal Loss (DFL) \cite{li2020generalized} and Complete IoU (CIoU) Loss \cite{zheng2021enhancing}, where DFL is used to expand the probability of the value around the object $y$. Its equation is shown as follows:
\begin{equation}
\label{dfl}
DFL(\mathcal{S}_{n}, \mathcal{S}_{n+1})=-((y_{n+1}-y) \log(\mathcal{S}_{n}) +(y-y_{n}) \log (\mathcal{S}_{n+1})),
\end{equation}
where the equations of $\mathcal{S}_n$ and $\mathcal{S}_{n+1}$ are shown below:
\begin{equation}
\mathcal{S}_{n}=\frac{y_{n+1}-y}{y_{n+1}-y_n}, \; \mathcal{S}_{n+1}=\frac{y-y_n}{y_{n+1}-y_n}.
\end{equation}
CIoU Loss adds an influence factor to Distance IoU (DIoU) Loss \cite{zheng2020distance} by considering the aspect ratio of the prediction and the ground truth bounding box. The equation is shown below:
\begin{equation}
CIoU_{Loss}=1-IoU+\frac{Distance_2^2}{Distance_C^2}+\frac{v^2}{(1-IoU)+\nu},
\end{equation}
where $\nu$ is the parameter that measures the consistency of the aspect ratio, defined as follows:
\begin{equation}
\nu=\frac{4}{\pi^2}(arctan\frac{w^{gt}}{h^{gt}}-arctan\frac{w^p}{h^p})^2,
\end{equation}
where $w$ is the weight of the bounding box, and $h$ is the height of the bounding box.

\section*{Experiments}
\subsection*{Dataset}
Medical University of Graz provides a public dataset named GRAZPEDWRI-DX \cite{nagy2022pediatric}, which consists of 20,327 X-ray images of wrist trauma in children. These images were collected from 6,091 patients between 2008 and 2018 by multiple pediatric radiologists at the Department of Pediatric Surgery of the University Hospital Graz. The images are annotated in 9 different classes by placing bounding boxes on them.

To perform the experiments shown in Table \ref{yolov8_640} and Table \ref{yolov8_1024}, we divide the GRAZPEDWRI-DX dataset randomly into three sets: training set, validation set, and test set. The sizes of these sets are approximately 70\%, 20\%, and 10\% of the original dataset, respectively. Specifically, our training set consists of 14,204 images (69.88\%), our validation set consists of 4,094 images (20.14\%), and our test set consists of 2,029 images (9.98\%). The code for splitting the dataset can be found on our GitHub. We also provide csv files of training, validation and test data on our GitHub, but it should be noted that each split is random and therefore not reproducible.

\subsection*{Evaluation Metric}
\subsubsection*{Intersection over Union (IoU)}
Intersection over Union (IoU) is a classical metric for evaluating the performance of the model for object detection. It calculates the ratio of the overlap and union between the generated candidate bounding box and the ground truth bounding box, which measures the intersection of these two bounding boxes. The IoU is represented by the following equation:
\begin{equation}
\label{IoU}
IoU=\frac{area(C)\cap area(G)}{area(C)\cup area(G)},
\end{equation}
where $C$ represents the generated candidate bounding box, and $G$ represents the ground truth bounding box containing the object. The performance of the model improves as the IoU value increases, with higher IoU values indicating less difference between the generated candidate and ground truth bounding boxes.

\subsubsection*{Precision-Recall Curve}
Precision-Recall Curve (P-R Curve) \cite{boyd2013area} is a curve with recall as the x-axis and precision as the y-axis. Each point represents a different threshold value, and all points are connected as a curve. The recall (R) and precision (P) are calculated according to the following equations:
\begin{equation}
Recall=\frac{T_P}{T_P+F_N}, \; Precision=\frac{T_P}{T_P+F_P},
\end{equation}
where True Positive ($T_P$) denotes the prediction result as a positive class and is judged to be true; False Positive ($F_P$) denotes the prediction result as a positive class but is judged to be false, and False Negative ($F_N$) denotes the prediction result as a negative class but is judged to be false.

\begin{table}[t]
\caption{Validation results of YOLOv8 for each class on the GRAZPEDWRI-DX dataset when the input image size is 1024.}
\label{class_yolov8}
\begin{tabular}{cccccccc}
\hline
\textbf{Class} & \textbf{Boxes} & \textbf{Instances} & \textbf{Precision} & \textbf{Recall} & \textbf{\begin{tabular}[c]{@{}c@{}} mAP$\rm ^{val}$\\ 50\end{tabular}} & \textbf{\begin{tabular}[c]{@{}c@{}}mAP$\rm ^{val}$\\ 50-95\end{tabular}} \\ \hline
all & 47435 & 9613 & 0.674 & 0.605 & 0.623 & 0.395 \\
boneanomaly & 276 & 53 & 0.505 & 0.094 & 0.110 & 0.035 \\
bonelesion & 45 & 8 & 0.629 & 0.250 & 0.416 & 0.212 \\
fracture & 18090 & 3740 & 0.885 & 0.903 & 0.947 & 0.572 \\
metal & 818 & 168 & 0.878 & 0.899 & 0.920 & 0.768 \\
periostealreaction & 3453 & 697 & 0.645 & 0.684 & 0.689 & 0.357 \\
pronatorsign & 567 & 104 & 0.561 & 0.713 & 0.611 & 0.338 \\
softtissue & 464 & 89 & 0.324 & 0.315 & 0.251 & 0.125 \\
text & 23722 & 4754 & 0.961 & 0.984 & 0.991 & 0.750 \\ \hline
\end{tabular}
\end{table}

\begin{table}[!t]
\caption{Validation results of our model for each class on the GRAZPEDWRI-DX dataset when the input image size is 1024.}
\label{class_ours}
\begin{tabular}{cccccccc}
\hline
\textbf{Class} & \textbf{Boxes} & \textbf{Instances} & \textbf{Precision} & \textbf{Recall} & \textbf{\begin{tabular}[c]{@{}c@{}} mAP$\rm ^{val}$\\ 50\end{tabular}} & \textbf{\begin{tabular}[c]{@{}c@{}}mAP$\rm ^{val}$\\ 50-95\end{tabular}} \\ \hline
all & 47435 & 9613 & 0.694 & 0.592 & 0.631 & 0.402 \\
boneanomaly & 276 & 53 & 0.510 & 0.151 & 0.169 & 0.076 \\
bonelesion & 45 & 8 & 0.658 & 0.243 & 0.414 & 0.213 \\
fracture & 18090 & 3740 & 0.899 & 0.896 & 0.947 & 0.569 \\
metal & 818 & 168 & 0.898 & 0.890 & 0.924 & 0.780 \\
periostealreaction & 3453 & 697 & 0.721 & 0.654 & 0.700 & 0.359 \\
pronatorsign & 567 & 104 & 0.534 & 0.683 & 0.611 & 0.342 \\
softtissue & 464 & 89 & 0.367 & 0.236 & 0.241 & 0.120 \\
text & 23722 & 4754 & 0.961 & 0.981 & 0.991 & 0.754 \\ \hline
\end{tabular}
\end{table}

\begin{table}[!t]
\caption{Model performance comparison of YOLOv8 models using SGD and Adam optimizers. For training with the SGD optimizer, the initial learning rate is 1×$10^{-2}$; for training with the Adam optimizer, the initial learning rate is 1×$10^{-3}$.}
\label{comparison}
\begin{tabular}{ccccccc}
\hline
\textbf{Model} & \textbf{Size} & \textbf{Optimizer} & \textbf{Best Epoch} & \textbf{\begin{tabular}[c]{@{}c@{}} mAP$\rm ^{val}$\\ 50\end{tabular}} & \textbf{\begin{tabular}[c]{@{}c@{}}mAP$\rm ^{val}$\\ 50-95\end{tabular}} & \textbf{\begin{tabular}[c]{@{}c@{}}Speed GPU\\ RTX 3080Ti\end{tabular}} \\ \hline
YOLOv8s & 640 & SGD & 56 & 0.611 & 0.389 & 4.4ms \\
YOLOv8s & 640 & Adam & 57 & 0.604 & 0.383 & 4.3ms \\
YOLOv8s & 1024 & SGD & 36 & 0.623 & 0.395 & 5.4ms \\
YOLOv8s & 1024 & Adam & 47 & 0.625 & 0.399 & 4.9ms \\
YOLOv8m & 640 & SGD & 52 & 0.621 & 0.396 & 4.9ms \\
YOLOv8m & 640 & Adam & 62 & 0.621 & 0.403 & 5.5ms \\
YOLOv8m & 1024 & SGD & 35 & 0.624 & 0.402 & 9.9ms \\
YOLOv8m & 1024 & Adam & 70 & 0.626 & 0.401 & 10.0ms \\ \hline
\end{tabular}
\end{table}

\begin{table}[ht]
\caption{Quantitative comparison of fracture detection when the input image size is 640. Speed means the total time of validate per image, and the total time includes the preprocessing, inference, and post-processing time.}
\label{yolov8_640}
\begin{tabular}{ccccccc}
\hline
\textbf{Model} & \textbf{\begin{tabular}[c]{@{}c@{}} mAP$\rm ^{val}$\\ 50\end{tabular}} & \textbf{\begin{tabular}[c]{@{}c@{}}mAP$\rm ^{val}$\\ 50-95\end{tabular}} & \textbf{\begin{tabular}[c]{@{}c@{}}Speed CPU\\ Intel Core i5\end{tabular}} & \textbf{\begin{tabular}[c]{@{}c@{}}Speed GPU\\ RTX 3080Ti\end{tabular}} & \textbf{PARAMS} & \textbf{FLOPs} \\ \hline
YOLOv5n & 0.589 & 0.339 & \textbackslash{} & 2.8ms & 1.77M & 4.2B \\
YOLOv8n & 0.601 & 0.374 & 67.4ms & 2.9ms & 3.01M & 8.1B \\
\textbf{Ours} & 0.605 & 0.379 & 111.3ms & 3.4ms & 3.01M & 8.2B \\ \hline
YOLOv5s & 0.601 & 0.357 & \textbackslash{} & 3.3ms & 7.03M & 15.8B \\
YOLOv8s & 0.604 & 0.383 & 191.5ms & 4.3ms & 11.13M & 28.5B \\
\textbf{Ours} & 0.612 & 0.392 & 285.1ms & 4.9ms & 11.13M & 28.7B \\ \hline
YOLOv5m & 0.613 & 0.371 & \textbackslash{} & 4.0ms & 20.89M & 48.0B \\
YOLOv8m & 0.621 & 0.403 & 536.4ms & 5.5ms & 25.84M & 78.7B \\
\textbf{Ours} & 0.629 & 0.404 & 685.9ms & 5.1ms & 25.84M & 78.7B \\ \hline
YOLOv5l & 0.620 & 0.379 & \textbackslash{} & 5.6ms & 46.15M & 107.8B \\
YOLOv8l & 0.624 & 0.403 & 1006.3ms & 7.4ms & 43.61M & 164.9B \\
\textbf{Ours} & 0.637 & 0.406 & 1370.8ms & 7.2ms & 43.61M & 164.9B \\ \hline
\end{tabular}
\end{table}

\begin{table}[!t]
\caption{Quantitative comparison of fracture detection when the input image size is 1024. Speed means the total time of validate per image, and the total time includes the preprocessing, inference, and post-processing time.}
\label{yolov8_1024}
\begin{tabular}{ccccccc}
\hline
\textbf{Model} & \textbf{\begin{tabular}[c]{@{}c@{}} mAP$\rm ^{val}$\\ 50\end{tabular}} & \textbf{\begin{tabular}[c]{@{}c@{}}mAP$\rm ^{val}$\\ 50-95\end{tabular}} & \textbf{\begin{tabular}[c]{@{}c@{}}Speed CPU\\ Intel Core i5\end{tabular}} & \textbf{\begin{tabular}[c]{@{}c@{}}Speed GPU\\ RTX 3080Ti\end{tabular}} & \textbf{PARAMS} & \textbf{FLOPs} \\ \hline
YOLOv5n & 0.600 & 0.347 & \textbackslash{} & 3.2ms & 1.77M & 4.2B \\
YOLOv8n & 0.605 & 0.387 & 212.1ms & 3.3ms & 3.01M & 8.1B \\
\textbf{Ours} & 0.608 & 0.391 & 260.4ms & 4.4ms & 3.01M & 8.1B \\ \hline
YOLOv5s & 0.622 & 0.371 & \textbackslash{} & 4.4ms & 7.03M & 15.8B \\
YOLOv8s & 0.625 & 0.399 & 519.5ms & 4.9ms & 11.13M & 28.5B \\
\textbf{Ours} & 0.631 & 0.402 & 717.1ms & 6.2ms & 11.13M & 28.5B \\ \hline
YOLOv5m & 0.624 & 0.380 & \textbackslash{} & 7.1ms & 20.89M & 48.0B \\
YOLOv8m & 0.626 & 0.401 & 1521.5ms & 10.0ms & 25.84M & 78.7B \\
\textbf{Ours} & 0.635 & 0.411 & 1724.4ms & 9.4ms & 25.85M & 78.7B \\ \hline
YOLOv5l & 0.626 & 0.378 & \textbackslash{} & 11.3ms & 46.15M & 107.8B \\
YOLOv8l & 0.636 & 0.404 & 2671.1ms & 15.1ms & 43.61M & 164.9B \\
\textbf{Ours} & 0.638 & 0.415 & 3864.5ms & 13.6ms & 43.61M & 164.9B \\ \hline
\end{tabular}
\end{table}

\begin{figure}[ht]
  \centering
  \includegraphics[width=\linewidth]{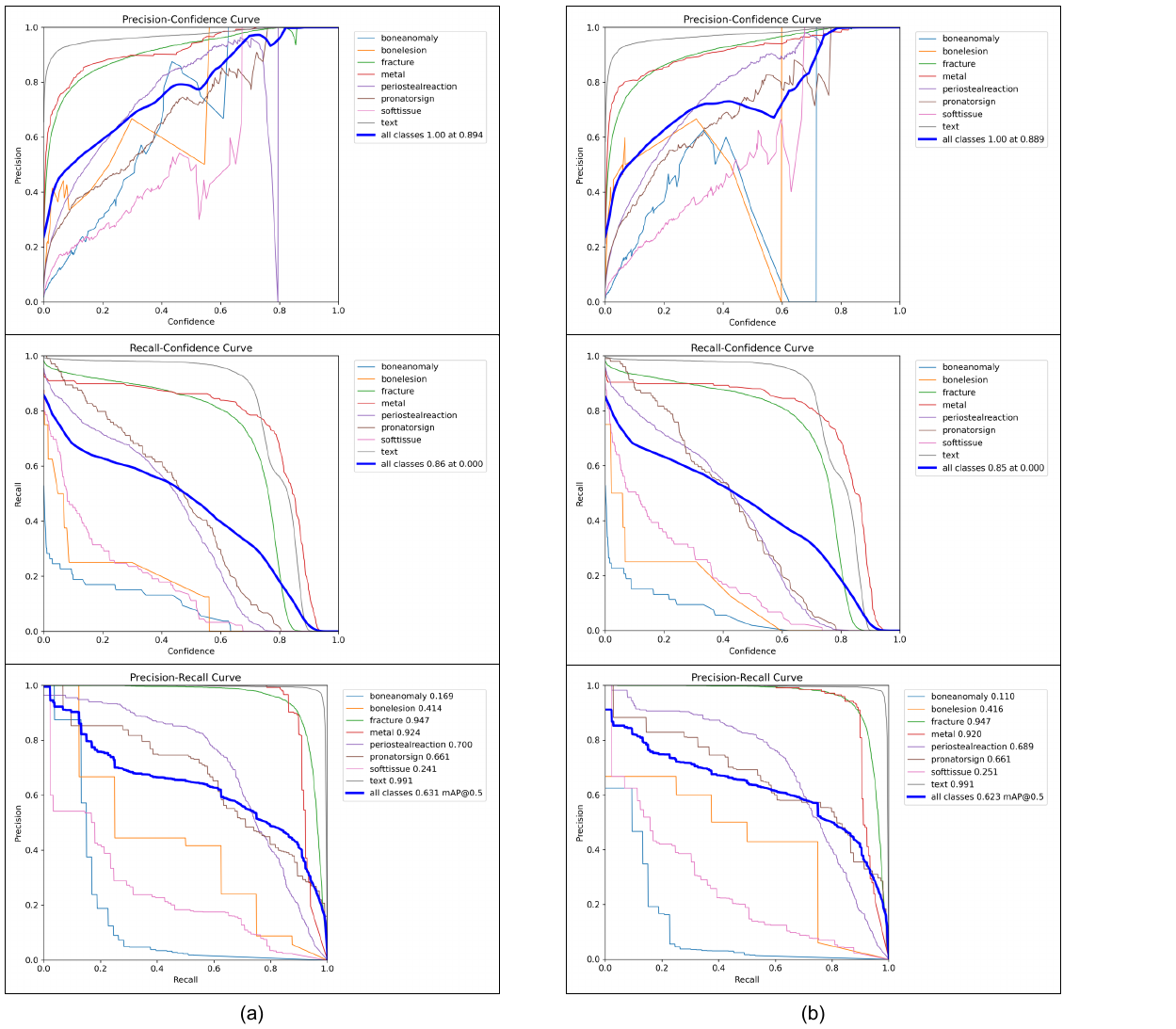}
  \caption{Detailed illustration of the validation at the input image size of 1024, (a) is our model, and (b) is YOLOv8 model.}
  \label{fig_curve}
\end{figure}

\begin{figure}[ht]
  \centering
  \includegraphics[width=\linewidth]{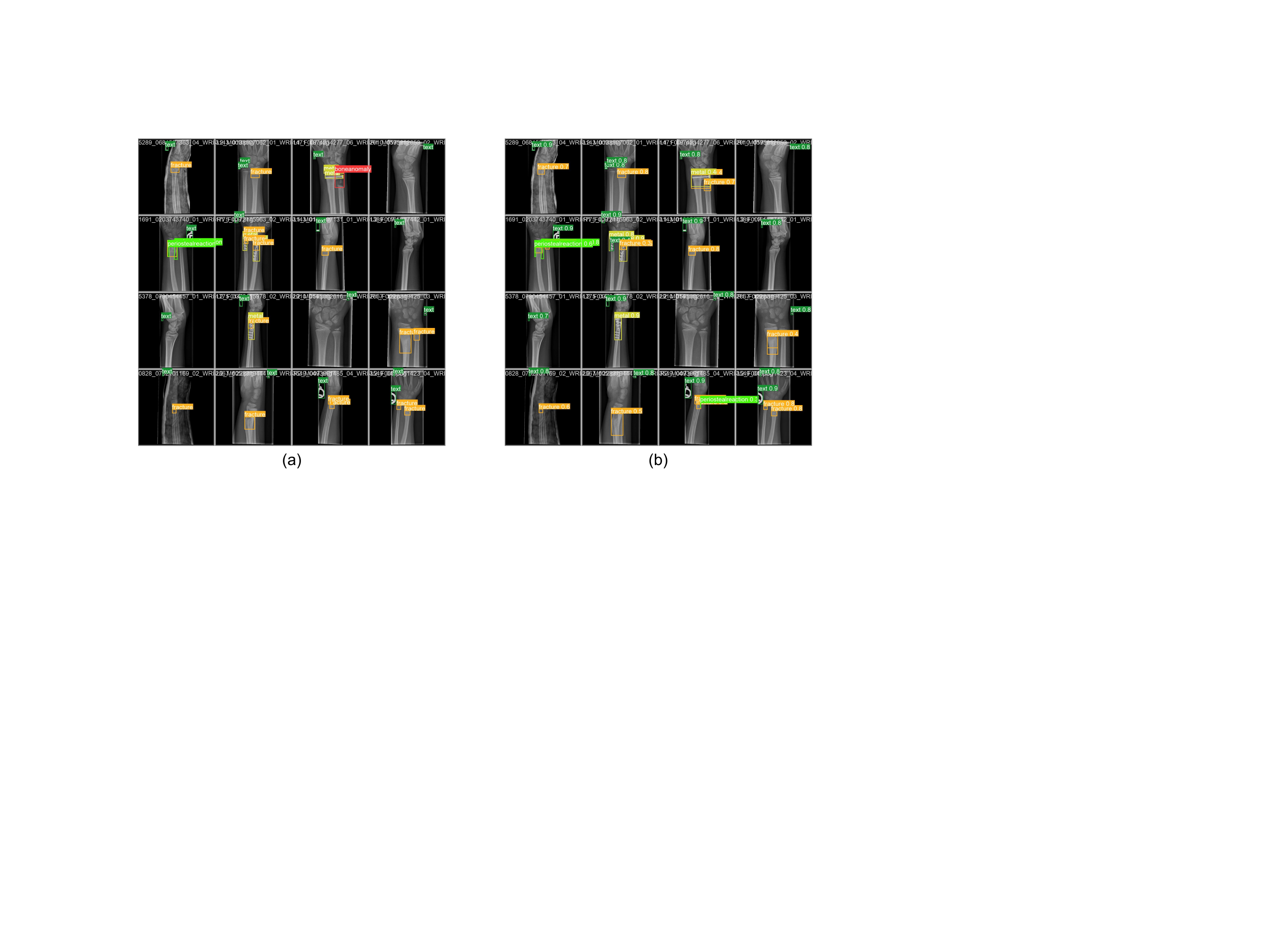}
  \caption{Examples of pediatric wrist fracture detection on X-ray images. (a) manually labeled images, (b) predicted images.}
  \label{fig_result}
\end{figure}

\begin{table}[ht]
\caption{Evaluation of wrist fracture detection with other state-of-the-art (SOTA) models on the GRAZPEDWRI-DX dataset.}
\label{sota}
\begin{tabular}{ccccc}
\hline
\textbf{Model} & \textbf{Precision} & \textbf{Recall} & \textbf{F1} & \textbf{\begin{tabular}[c]{@{}c@{}} mAP$\rm ^{val}$\\ 50\end{tabular}} \\ \hline
YOLOv5\cite{glenn2022} & 0.682 & 0.581 & 0.607 & 0.626 \\
YOLOv7\cite{wang2022yolov7} & 0.556 & 0.582 & 0.569 & 0.628 \\
YOLOv7\cite{wang2022yolov7} + CBAM\cite{woo2018cbam} & 0.709 & 0.593 & 0.646 & 0.633 \\
YOLOv7\cite{wang2022yolov7} + GAM\cite{liu2021global} & 0.745 & 0.574 & 0.646 & 0.634 \\
YOLOv8\cite{glenn2023} & 0.694 & 0.679 & 0.623 & 0.636 \\
\textbf{Ours} & 0.734 & 0.592 & 0.635 & 0.638 \\ \hline
\end{tabular}
\end{table}

\begin{figure}[ht]
  \centering
  \includegraphics[width=\linewidth]{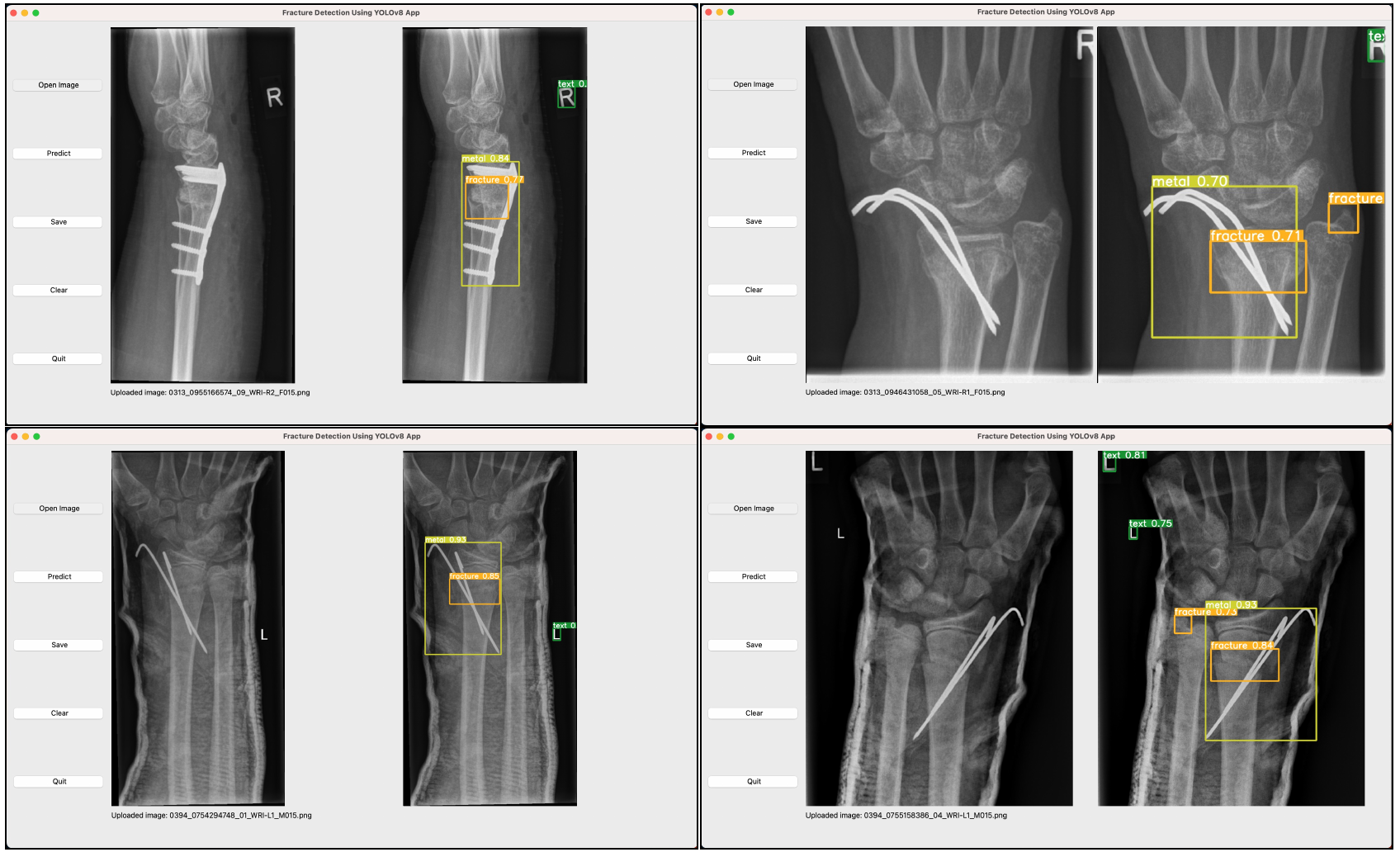}
  \caption{Example of using the application ``Fracture Detection with YOLOv8 Application'' on macOS operating system .}
  \label{fig_app}
\end{figure}

\subsubsection*{F1-score}
The F-score is a commonly used metric to evaluate the model accuracy, providing a balanced measure of performance by incorporating both precision and recall. The F-score equation is as follows:
\begin{equation}
\label{f-score}
\operatorname{\emph{F-score}} = \frac{\left(1+\beta^{2}\right) \times Precision \times Recall}{\beta^{2} \times Precision + Recall}
\end{equation}
When $\beta$ = 1, the F1-score is determined by the harmonic mean of precision and recall, and its equation is as follows:
\begin{equation}
\label{f1-score}
F_{1}=\frac{2 \times Precision \times Recall}{Precision + Recall}=\frac{2T_P}{2T_P+F_P+F_N}
\end{equation}

\subsection*{Experiment Setup}
During the model training process, we utilize pre-trained YOLOv8 model from the MS COCO (Microsoft Common Objects in Context) val2017 dataset \cite{lin2014microsoft}. The research reports provided by Ultralytics \cite{glenn2022,glenn2023} suggests that YOLOv5 training requires 300 epochs, while training YOLOv8 requires 500 epochs. Since we use pre-trained model, we initially set the total number of epochs to 200 with a patience of 50, which indicate that the training would end early if no observable improvement is noticed after waiting for 50 epochs. In the experiment comparing the effect of the optimizer on the model performance, we notice that the best epcoh of all the models is within 100, as shown in Table \ref{comparison}, mostly concentrated between 50 and 70 epochs. Therefore, to save computing resources, we adjust the number of epochs for our model training to 100.

As the suggestion \cite{glenn2023} of Glenn, for model training hyperparameters, the Adam \cite{kingma2014adam} optimizer is more suitable for small custom datasets, while the SGD \cite{ruder2016overview} optimizer perform better on larger datasets. To prove the above conclusion, we train YOLOv8 algorithm models using the Adam and SGD optimizers, respectively, and compare the effects on the model performance. The comparison results are shown in Table \ref{comparison}.

For the experiments, we choose the SGD optimizer with an initial learning rate of 1$\times10^{-2}$, a weight decay of 5$\times10^{-4}$, and a momentum of 0.937 during our model training. We set the input image size to 640 and 1024 for training on a single GPU GeForce RTX 3080Ti 12GB with a batch size of 16. We train the model using Python 3.8 and PyTorch 1.8.2, and recommend readers to use Python 3.7 or higher and PyTorch 1.7 or higher for training. It is noteworthy that due to GPU memory limitations, we choose 3 worker threads to load data on GPU GeForce RTX 3080Ti 12GB when training our model. Therefore, using GPUs with larger memory and more computing power can effectively increase the speed of model training.

\subsection*{Ablation Study}
In order to demonstrate the positive effect of our training method on the performance of YOLOv8 model, we conduct an ablation study on YOLOv8s model by calculating each evaluation metric for each class, as shown in Table \ref{class_yolov8}. Among all classes, YOLOv8s model has good accuracy in detecting fracture, metal and text, with mAP 50 of each above 0.9. On the opposite, the detection ability of bone-anomaly is poor, with mAP 50 of 0.11. Therefore, we increase the contrast and brightness of X-ray images to make bone-anomaly easier to detect. Table \ref{class_ours} presents the predictions of YOLOv8s model using our training method for each class. Compared with YOLOv8s model, the mAP value predicted by the model using our training method for bone-anomaly increased from 0.11 to 0.169, an increase of 53.6\%. Figure \ref{fig_curve} also shows that our model has a better performance in detecting bone-anomaly, which enables the improvement of the overall model performance. From the ablation study presented above, we demonstrate that the model performance can be improved by using our training method (data augmentation). In addition to the data enhancement, researchers can also improve model performance by adding modules such as the Convolutional Block Attention Module (CBAM) \cite{woo2018cbam}.

\subsection*{Experimental Results}
\label{experiment}
Before training our model, in order to choose an optimizer that has a more positive effect on the model performance, we compare the performance of models trained with the SGD \cite{ruder2016overview} optimizer and the Adam \cite{kingma2014adam} optimizer. As shown in Table \ref{comparison}, using the SGD optimizer to train the model requires less epochs of weight updates. Specifically, for YOLOv8m model with an input image size of 1024, the model trained with the SGD optimizer achieves the best performance at the 35th epoch, while the best performance of the model trained with the Adam optimizer is at the 70th epoch. In terms of mAP and inference time, there is not much difference in the performance of the models trained with the two optimizers. Specifically, when the input image size is 640, the mAP value of YOLOv8s model trained with the SGD optimizer is 0.007 higher than that of the model trained with the Adam optimizer, while the inference time is 0.1ms slower. Therefore, according to the above experimental results and the suggestion by Glenn \cite{glenn2022,glenn2023}, for YOLOv8 model training on a training set of 14,204 X-ray images, we choose the Adam optimizer. However, after using data augmentation, the number of X-ray images in the training set extend to 28,408, so we switch to the SGD optimizer to train our model.

After using data augmentation, our models have a better mAP value than that of YOLOv8 model, as shown in Table \ref{yolov8_640} and Table \ref{yolov8_1024}. Specifically, when the input image size is 640, compared with YOLOv8m model and YOLOv8l model, the mAP 50 of our model improves from 0.621 to 0.629, and from 0.623 to 0.637, respectively. Although the inference time on the CPU is increased from 536.4ms and 1006.3ms to 685.9ms and 1370.8ms, respectively, the number of parameters and FLOPs are the same, which means that our model can be deployed on the same computing power platform. In addition, we compare the performance of our model with that of YOLOv7 and its improved models. As shown in Table \ref{sota}, the mAP value of our model is higher than those of YOLOv7 \cite{wang2022yolov7}, YOLOv7 with Convolutional Block Attention Module (CBAM) \cite{woo2018cbam} and YOLOv7 model with Global Attention Mechanism (GAM) \cite{liu2021global}, which demonstrates that our model has obtained SOTA performance.

This paper aims to design a pediatric wrist fracture detection application, so we use our model for fracture detection. Figure \ref{fig_result} shows the results of manual annotation by the radiologist and the results predicted using our model. These results demonstrate that our model has a good ability to detect fractures in single fracture cases, but metal puncture and dense multiple fracture situations badly affects the accuracy of prediction.

\section*{Application}
\label{application}
After completing model training, we utilize a Python library that includes the Qt toolkit, PySide6, to develop a Graphical User Interface (GUI) application. Specifically, PySide6 is the Qt6-based version of the PySide GUI library from the Qt Company.

According to the model performance evaluation results in Table \ref{yolov8_640} and Table \ref{yolov8_1024}, we choose our model with YOLOv8s algorithm and the input image size of 1024, to perform fracture detection. Our model is exported to onnx format, and is applied to the GUI application. Figure \ref{fig_app} depicts the flowchart of the GUI application operation on macOS. As can be seen from the illustration, our application is named ``Fracture Detection Using YOLOv8 App''. Users can open and predict the images, and save the predictions in this application. In summary, our application is designed to assist pediatric surgeons in analyzing fractures on pediatric wrist trauma X-ray images.

\section*{Conclusions and Future Work}
\label{conclusion}
Ultralytics proposed the latest version of YOLO series (YOLOv8) in 2023. Although there are relatively few research works on YOLOv8 model for medical image processing, we apply it to fracture detection and use data augmentation to improve the model performance. We randomly divide the dataset, consisting of 20,327 pediatric wrist trauma X-ray images from 6,091 patients, into training, test, and validation sets to train the model and evaluate the performance.

Furthermore, we develop an application named "Fracture Detection Using YOLOv8 App" to analyze pediatric wrist trauma X-ray images for fracture detection. Our application aims to assist pediatric surgeons in interpreting X-ray images, reduce the probability of misclassification, and provide a better information base for surgery. The application is currently available for macOS, and in the future, we plan to deploy different sizes of our model in the application, and extend the application to iOS and Android. This will enable inexperienced pediatric surgeons in hospitals located in underdeveloped areas to use their mobile devices to analyze pediatric wrist X-ray images.

In addition, we provide the specific steps for training the model and the trained model in our GitHub. If readers wish to use YOLOv8 model to detect fracture in other parts of the body except the pediatric wrist, they can use our trained model as the pre-training model, which can greatly improve the performance of the model.

\bibliography{sample}

\section*{Author contributions statement}
R.J. conceived and conducted the experiments, and W.C. provided knowledge about the fracture and analyzed the results. All authors have reviewed the manuscript. 

\section*{Funding}
The authors did not receive support from any organization for the submitted work.

\section*{Competing interests}
The authors have no financial or proprietary interests in any material discussed in this article.

\section*{Ethics approval}
This research does not involve human participants and/or animals.

\section*{Data availability}
The datasets analysed during the current study are available at Figshare under \url{https://doi.org/10.6084/m9.figshare.14825193.v2}. The implementation code and the trained model for this study can be found on GitHub at \url{https://github.com/RuiyangJu/Bone_Fracture_Detection_YOLOv8}.

\end{document}